\documentclass[sigconf]{acmart}

\usepackage{enumitem}
\usepackage{colortbl,gensymb,multirow}
\usepackage{listings}
\usepackage{xcolor}
\usepackage{booktabs}
\usepackage{multicol}
\usepackage{soul}
\usepackage{pifont}
\usepackage{afterpage}
\usepackage{placeins}

\definecolor{HeatRed}{HTML}{C9432F}
\definecolor{CanopyGreen}{HTML}{2F7D4E}
\definecolor{SkyBlue}{HTML}{2F6F9F}
\definecolor{WarmGray}{HTML}{6F625B}
\definecolor{SoftHeat}{HTML}{F7E1D5}
\definecolor{SoftCool}{HTML}{E4EEF6}
\newcommand{\yes}{\textcolor{CanopyGreen}{\ding{51}}}
\newcommand{\no}{\textcolor{HeatRed}{\ding{55}}}
\newcommand{\best}[1]{\textbf{#1}}

\lstdefinestyle{pythonstyle}{
    language=Python,
    backgroundcolor=\color{gray!8},
    basicstyle=\ttfamily\scriptsize,
    keywordstyle=\color{HeatRed}\bfseries,
    stringstyle=\color{CanopyGreen},
    commentstyle=\color{WarmGray}\itshape,
    numberstyle=\tiny\color{WarmGray},
    showstringspaces=false,
    frame=single,
    framesep=4pt,
    framerule=0.3pt,
    rulecolor=\color{gray!35},
    breaklines=true,
    breakatwhitespace=false,
    columns=flexible,
    keepspaces=true,
    upquote=true,
    abovecaptionskip=4pt,
    belowcaptionskip=4pt,
    postbreak=\mbox{\textcolor{gray}{$\hookrightarrow$}\space},
    morekeywords={as, from, with, in}
}
\lstdefinestyle{bashstyle}{
    language=bash,
    backgroundcolor=\color{gray!8},
    basicstyle=\ttfamily\scriptsize,
    keywordstyle=\color{HeatRed}\bfseries,
    commentstyle=\color{WarmGray}\itshape,
    frame=single,
    framesep=4pt,
    framerule=0.3pt,
    rulecolor=\color{gray!35},
    breaklines=true,
    columns=flexible,
    keepspaces=true,
    morekeywords={huggingface-cli, pip, python}
}
\AtBeginDocument{%
  }

\renewcommand\footnotetextcopyrightpermission[1]{}
\setcopyright{none}
\copyrightyear{2026}
\acmYear{2026}
\acmDOI{}
\acmConference[SIGSPATIAL '26]{Proceedings of the 34th ACM International Conference on Advances in Geographic Information Systems}{November 3--6, 2026}{Riverside, CA, USA}
\acmISBN{}

\begin{document}

\title{HeatCast: A Benchmark for Neighborhood-Scale LST\texorpdfstring{\\}{ }Forecasting across 124 U.S. Cities}

\author{Jesus Guerrero}
\orcid{0000-0001-9168-3414}
\affiliation{%
  \institution{University of Texas at San Antonio}
  \city{San Antonio}
  \state{TX}
  \country{USA}
}
\email{Jesus.Guerrero@utsa.edu}

\author{Isaac Corley}
\orcid{0000-0002-9273-7303}
\affiliation{%
  \institution{Taylor Geospatial}
  \city{San Antonio}
  \state{TX}
  \country{USA}
}
\email{isaac.corley@taylorgeospatial.org}

\author{Leon Najafirad}
\affiliation{%
  \institution{Independent Researcher}
  \city{San Antonio}
  \state{TX}
  \country{USA}
}
\email{Najafiradleon@gmail.com}

\author{Maryam Tabar}
\orcid{0009-0005-8492-1310}
\affiliation{%
  \institution{University of Texas at San Antonio}
  \city{San Antonio}
  \state{TX}
  \country{USA}
}
\email{Maryam.Tabar@utsa.edu}

\author{Paul Rad}
\orcid{0000-0001-9671-577X}
\affiliation{%
  \institution{University of Texas at San Antonio}
  \city{San Antonio}
  \state{TX}
  \country{USA}
}
\email{paul.rad@utsa.edu}

\authornote{Corresponding Author.}

\begin{abstract}
Land Surface Temperature (LST) is a widely used satellite-derived measure of urban surface heat, but there is no shared benchmark for forecasting it at 30~m. Prior studies usually cover one to three cities, use kilometer-scale products, or do not release data and code. We introduce \textbf{HeatCast}, a Landsat-based benchmark for monthly LST forecasting across 124 U.S. cities from 2013 through June~2025. HeatCast contains 30~m monthly tiles with LST, elevation, surface-reflectance RGB, three spectral indices, broadband albedo, quality masks, and Local Climate Zone (LCZ) labels, together with a fixed temporal split, LCZ-stratified metrics, and a reference evaluation harness. We evaluate a CNN+LSTM and Earthformer on next-month forecasting, where Earthformer reaches 7.74~K RMSE against 10.42~K for the CNN+LSTM. Forecasting from the eight non-LST channels alone reaches 7.72~K, against 8.15~K from LST history and 8.68~K from RGB. The data, code, and weights are released under MIT at \url{https://doi.org/10.57967/hf/9889}.
\end{abstract}

\begin{teaserfigure}
    \centering
    \includegraphics[width=0.88\textwidth]{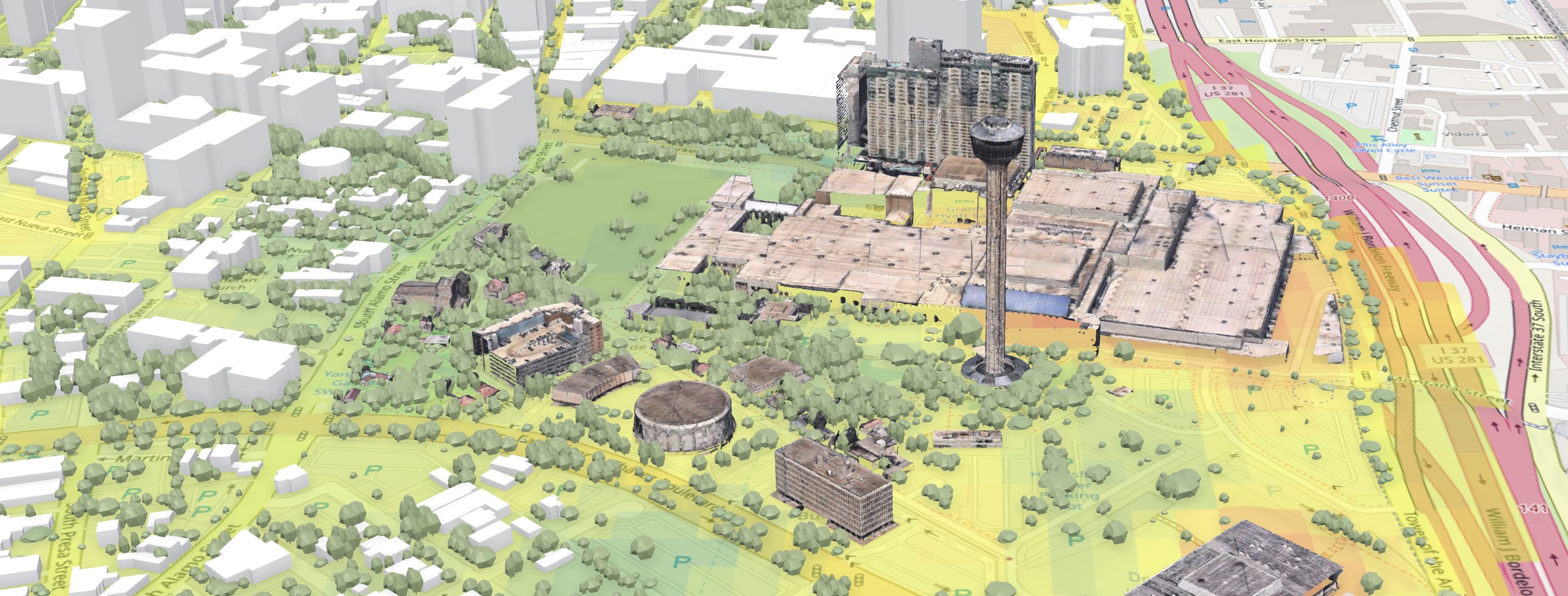}
    \caption{Landsat-derived LST over downtown San Antonio, TX. Surface temperature varies by 10--15~K within a single neighborhood. The warmest pixels occur over impervious surfaces and the coolest over vegetated parks and water. The benchmark evaluates next-month LST forecasts at this 30~m resolution.}
    \Description{A Landsat-derived land-surface-temperature heatmap over downtown San Antonio, Texas, at 30~m resolution, with warmer reds over impervious built-up surfaces and cooler blues over vegetated parks and water bodies.}
    \label{fig:main}
\end{teaserfigure}

\ccsdesc[500]{Applied computing~Earth and atmospheric sciences}
\ccsdesc[300]{Computing methodologies~Neural networks}
\ccsdesc[300]{Computing methodologies~Computer vision}

\keywords{Land Surface Temperature, Urban Heat Island, Benchmark Dataset, Spatiotemporal Forecasting, Remote Sensing, Climate Resilience}

\maketitle
\hypersetup{
  pdftitle={HeatCast: A Benchmark for Neighborhood-Scale LST Forecasting across 124 U.S. Cities},
  pdfauthor={Jesus Guerrero, Isaac Corley, Leon Najafirad, Maryam Tabar, Paul Rad}
}

\begin{table*}[!t]
\centering
\begin{minipage}{\textwidth}
\resizebox{\textwidth}{!}{
\begin{tabular}{@{}p{3.2cm}ccccp{2.2cm}p{1.5cm}l@{}}
\toprule
Paper & Baseline & Code & Data & Model & Availability & Temporal & Spatial \\
\midrule
Xu et al.~\cite{xu_novel_2024} & \yes & \no & \no & CNN & Indirect & None & 30~m input; in-situ output \\
\midrule
Ming et al.~\cite{ming_unraveling_2024} & \yes & \no & \no & GWR/RF & Upon request & Seasonal & 10~m--1~km input; 1~km output \\
\midrule
Yang et al.~\cite{yang_global_2024} & \no & \no & \yes & Statistical & Link in paper & Monthly & 9~km \\
\midrule
Tehrani et al.~\cite{tehrani_predicting_2024} & \yes & \no & \no & GRU/ANN & Upon request & Annual & City-level \\
\midrule
Lauwaet et al.~\cite{lauwaet_high_2024} & \yes & \no & \no & Physics & Upon request & Hourly & 100~m \\
\midrule
Han et al.~\cite{han_time-continuous_2024} & \yes & \no & \no & U-Net++ & Unavailable & Daily & $0.1^\circ$ ($\sim$10~km) \\
\midrule
Liu et al.~\cite{liu2025daily} & \yes & \no & \no & Ensemble DL & Unavailable & Daily & 100~m \\
\midrule
\rowcolor{SoftHeat}
\textbf{HeatCast (Ours)} & \yes & \yes & \yes & Transformers & Hugging Face & Monthly & \textbf{30~m} \\
\bottomrule
\end{tabular}
}
\end{minipage}
\caption{LST-related datasets and methods covering more than ten urban areas. Prior work focuses on retrieval, gap filling, or UHI characterization. HeatCast instead evaluates LST \textit{forecasting} at 30~m across 124 U.S. cities.}
\label{tab:comparison}
\end{table*}

\section{Introduction}
High ambient land-surface temperatures (LST) are associated with excess
mortality~\cite{gasparrini2015mortality}. Within a single city, LST varies sharply over short distances, since impervious surfaces run substantially warmer than vegetation and water, producing block-scale thermal gradients that track patterns of urban development (Figure~\ref{fig:main}). Lower-income and historically redlined neighborhoods are exposed to the hottest of these gradients~\cite{hsu2021disproportionate,hoffman2020effects}. Kilometer-scale LST products average over this structure, whereas Landsat channels resolve the individual blocks and land-cover features that urban heat planning acts on~\cite{akbari2009cool,calhoun2024estimating}.

Despite extensive data-driven LST modeling work~\cite{wang_machine_2023, ghorbany2024towards}, there is no widely adopted open benchmark for urban LST forecasting comparable to those used for land-cover mapping~\cite{stewart2023ssl4eo} or medium-range weather~\cite{rasp2020weatherbench, nathaniel2024chaosbench}. This makes model performance, auxiliary inputs, and task difficulty hard to compare across studies. Furthermore, existing studies are typically limited in geographic scope, spatial resolution, or reproducibility, as discussed in Section 2.

\begin{figure*}[t]
\centering
\includegraphics[width=0.95\textwidth]{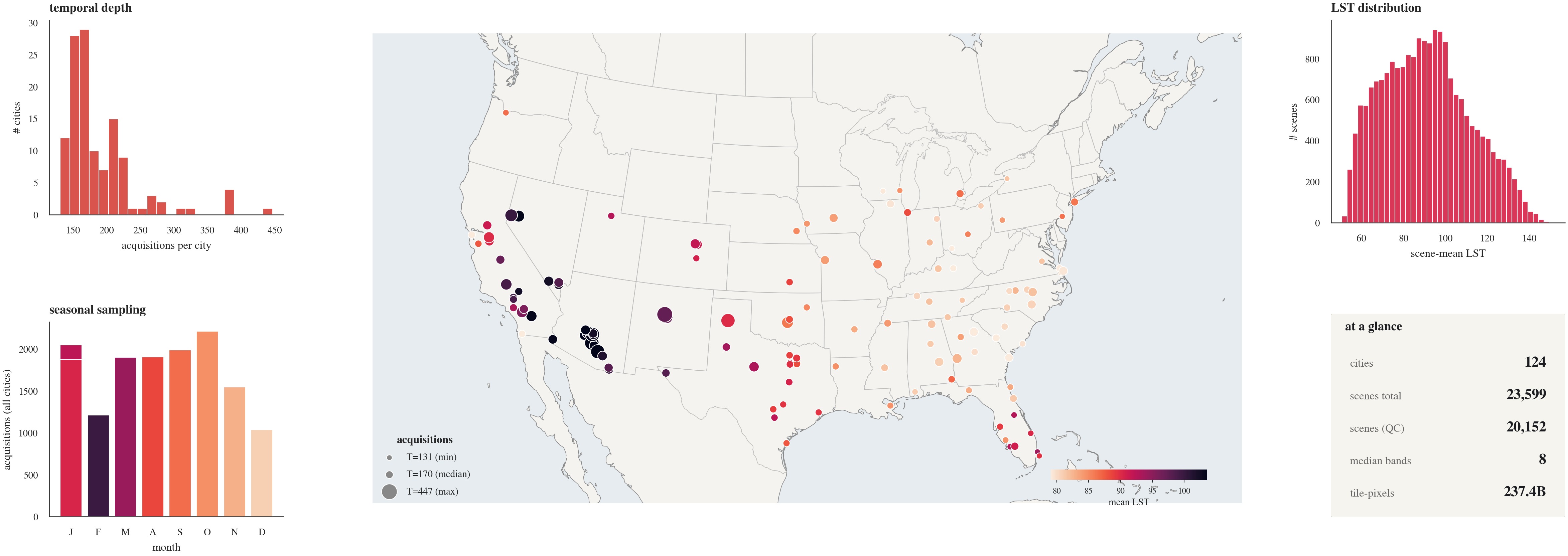}
\caption{\textbf{HeatCast covers 124 U.S. cities.} Marker color denotes per-city mean LST, and marker size denotes the number of acquisitions. The surrounding plots summarize temporal coverage, seasonal sampling, and scene-mean LST.}
\Description{Five-panel dataset overview: an Albers map of the contiguous United States with 124 city markers colored by mean LST and sized by acquisition depth, with histograms for per-city acquisitions and monthly cadence on the left, a pooled LST distribution histogram and a numerical at-a-glance stats card on the right.}
\label{fig:usa_image}
\end{figure*}

We introduce \textbf{HeatCast}, a benchmark for monthly LST forecasting on a 30~m grid across 124 U.S. cities and 12 years of Landsat~8/9 observations (Figure~\ref{fig:usa_image}). HeatCast defines a fixed next-month task, a temporal train/validation/test split, two reference baselines, and evaluation code that reports errors in Kelvin.

\noindent\textbf{Contributions.}
\begin{itemize}[leftmargin=*, topsep=2pt, itemsep=1pt]
    \item \textbf{Dataset.} HeatCast contains approximately 1.4~M $128\times128$ patches from 124 U.S. cities on a common 30~m grid. Each patch includes nine aligned channels and LCZ labels. The $\sim$150~GB dataset is released under MIT.
    \item \textbf{Benchmark protocol.} We define a next-month forecasting task with a 12-month input window, a temporal split (train 2013--2021, validation 2022--2023, test January~2024--June~2025), and LCZ-stratified evaluation in Kelvin. We distribute the exact samples as fixed manifests with a reference evaluator.
    \item \textbf{Baselines.} We train a CNN+LSTM and Earthformer on the temporal split. Earthformer obtains 7.74~K aggregate RMSE, compared with 10.42~K for the CNN+LSTM.
    \item \textbf{Feature-set ablation.} Earthformer achieves its lowest RMSE with the eight auxiliary channels (7.72~K), compared with all nine channels (7.74~K) and historical LST alone (8.15~K).
\end{itemize}

\section{Related Work}
\label{sec:related}

\noindent\textbf{LST datasets and forecasting studies.}
Most LST modeling studies cover one to three cities and use statistical or shallow-learning models~\cite{calhoun2024estimating, GWRSanAntonio, wen2024comparative, wang2025impact, ullah_impact_2025}. Their performance across climates is generally not tested. Continental and global studies typically use 1--10~km data~\cite{ming_unraveling_2024, yang_global_2024, niu_development_2021}, which effectively smooths intra-city temperature variation. Other studies retrieve LST from raw radiances, fuse Landsat and MODIS for gap filling~\cite{gao2006starfm, bouaziz2026deep}, or characterize UHIs statistically~\cite{bechtel2019suhi}. These tasks differ from forecasting LST from a multichannel observation history. Table~\ref{tab:comparison} compares LST-related work covering more than ten urban areas. Many of these datasets cannot serve as direct benchmarks because their code or data are unavailable, their ancillary inputs are region-specific, or their resolution is too coarse for neighborhood-scale evaluation.

\noindent\textbf{Geospatial and weather forecasting benchmarks.}
Adjacent areas of geospatial ML have open benchmarks with frozen splits. WeatherBench~\cite{rasp2020weatherbench} standardizes medium-range weather forecasting, ChaosBench~\cite{nathaniel2024chaosbench} extends evaluation to subseasonal-to-seasonal scales, EarthNet2021~\cite{requena2021earthnet2021} benchmarks Earth-surface forecasting from Sentinel-2, and SSL4EO-L~\cite{stewart2023ssl4eo} provides a Landsat pretraining benchmark. None target urban LST forecasting at 30~m.

\noindent\textbf{Spatio-temporal forecasting models and foundation models.}
Spatio-temporal forecasters such as ConvLSTM~\cite{shi2015convlstm}, PredRNN~\cite{wang2022predrnn}, SimVP~\cite{gao2022simvp}, and Earthformer~\cite{gao2022earthformer} are relevant to HeatCast because they model video or radar-style sequences. Geospatial foundation models such as SatMAE~\cite{cong2022satmae}, SSL4EO-L~\cite{stewart2023ssl4eo}, and the Prithvi family~\cite{jakubik2023prithvi} are a second relevant family. These models pretrain on raw spectral bands but have not been evaluated on thermal forecasting. We evaluate a CNN+LSTM and Earthformer as reference models.

\section{Dataset Construction}
\label{sec:dataset}

\subsection{Acquisition Pipeline}
We use 124 U.S. cities with Esri urban footprints larger than 90~mi$^2$. For each city, we query Landsat~8/9 Collection~2 Level-2 Surface Reflectance and Surface Temperature scenes from the Microsoft Planetary Computer STAC catalog. The archive spans mid-2013, following Landsat~8's commissioning, through June~2025. We select the lowest-cloud-cover scene per month using the Landsat quality-assessment band (QA\_PIXEL) and try up to twenty fallback scenes before marking a month as missing.

Following \citet{stewart2023ssl4eo} and \citet{corley2025landsat}, we mosaic scenes over each city footprint and reproject them to a common 30~m UTM grid. Surface reflectance values come from the LaSRC algorithm~\cite{vermote2018lasrc}. Surface temperature comes from the USGS Collection~2 single-channel ST product~\cite{malakar2018landsat}. USGS derives this product from the 100~m Thermal Infrared Sensor (TIRS) band and distributes it on the 30~m optical grid. Table~\ref{tab:data_sources} summarizes the data sources.

\begin{table}[t]
\centering
\caption{Data sources used to build HeatCast. All sources are public and have stable programmatic access paths.}
\label{tab:data_sources}
\resizebox{0.9\columnwidth}{!}{%
\begin{tabular}{@{}lll@{}}
\toprule
\textbf{Source} & \textbf{Data} & \textbf{Resolution} \\
\midrule
Landsat 8/9 C2L2 & Surface reflectance & 30~m \\
Landsat 8/9 TIRS & Surface temperature & 100~m $\rightarrow$ 30~m \\
NASADEM & Elevation & 30~m \\
CONUS LCZ & Climate zones & 100~m \\
Esri & Urban footprints & Vector \\
\bottomrule
\end{tabular}}
\end{table}

\subsection{Channels}
\begin{figure*}[t]
\centering
\includegraphics[width=0.93\textwidth]{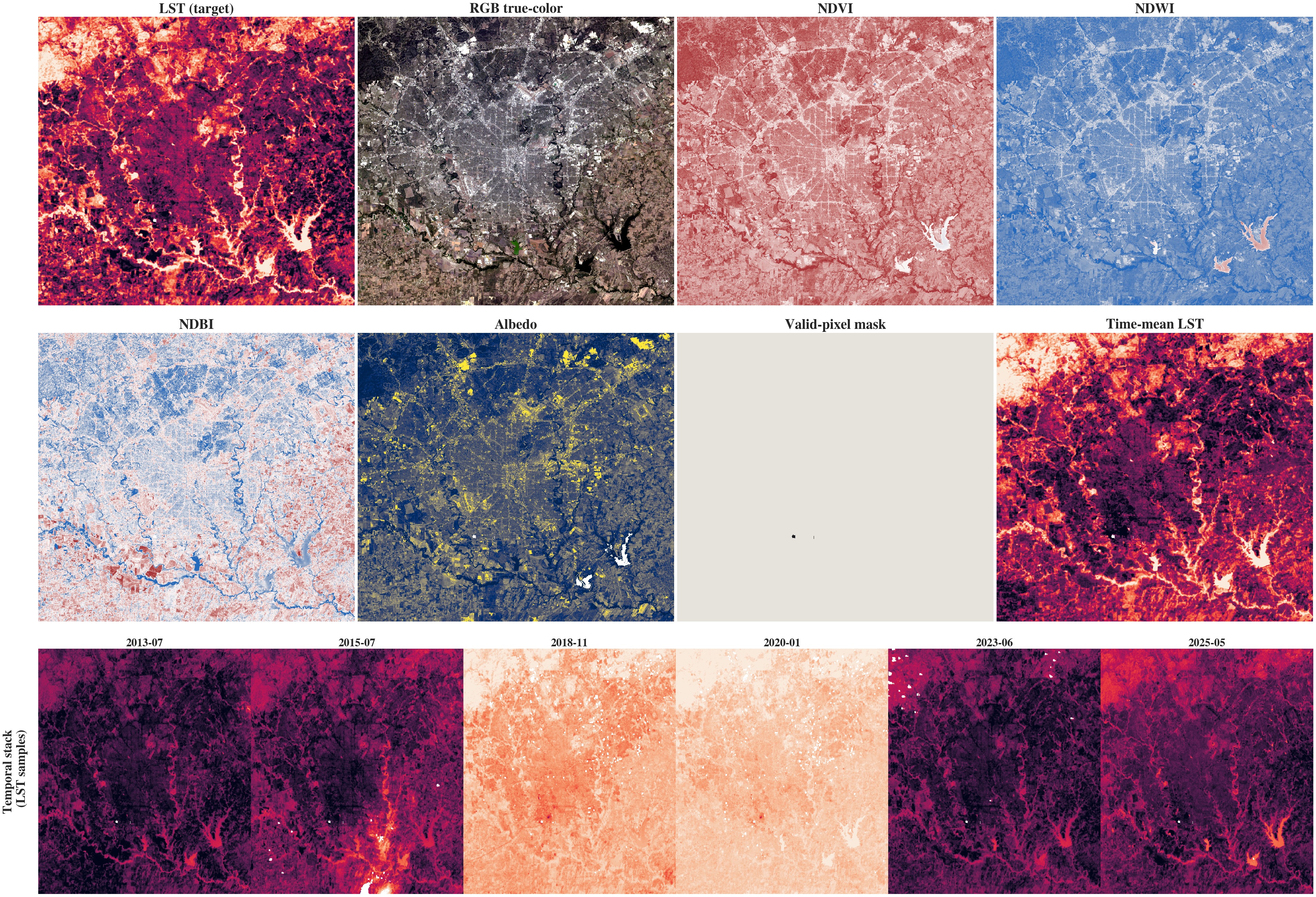}
\caption{\textbf{All HeatCast channels share the same 30~m grid.} The panels show LST, RGB, three spectral indices, broadband albedo, the valid-pixel mask, and time-mean LST over San Antonio, TX. The bottom row shows six monthly LST observations of the same tile.}
\Description{An eight-panel mosaic showing one San Antonio scene rendered across every HeatCast channel, plus a strip of six monthly LST acquisitions along the bottom.}
\label{fig:channel_mosaic}
\end{figure*}

Each tile has nine input channels per month: LST, a digital elevation model (DEM), surface-reflectance RGB (bands~2--4), three spectral indices (NDVI~\cite{tucker1979ndvi}, NDWI~\cite{mcfeeters1996use}, and NDBI~\cite{zha2003use}), and Liang broadband albedo~\cite{liang2001narrowband}. Figure~\ref{fig:channel_mosaic} shows one tile across all nine channels. We compute the indices with Equations~\ref{ndvieq}--\ref{NDBIeq}, using the open-water form of NDWI from \citet{mcfeeters1996use}. We compute albedo as the linear band combination $\alpha = c_0 + \sum_i c_i \rho_{B_i}$ from \citet{liang2001narrowband}.

Prior work reports associations between LST and each auxiliary variable. NDVI and NDWI describe vegetation and water~\cite{weng2004estimation, roy2022examining}. NDBI and albedo describe built surfaces and reflectance~\cite{chen2006remote, tahooni2023monitoring} while DEM provides elevation~\cite{he2019impact}. Table~\ref{tab:channels} lists the storage encoding and observed range of each channel.

\begin{table}[t]
\centering
\small
\caption{Channel specifications and storage ranges. Each channel uses int16 storage with a fixed linear scale. DEM carries a +10,000 offset for negative elevations, and LST is converted from $^\circ$C to Kelvin for evaluation. The extrema include residual retrieval outliers, which are masked as NoData before evaluation.}
\label{tab:channels}
\begin{tabular}{llll}
\toprule
\textbf{Channel} & \textbf{Min} & \textbf{Max} & \textbf{Units} \\
\midrule
DEM & $-$101 & 3,110 & m (stored int16 + 10k) \\
LST & $-$123 & 99 & $^\circ$C (stored int16, eval in K) \\
Red & 1 & 10,000 & DN ($\times$10k) \\
Green & 1 & 10,000 & DN ($\times$10k) \\
Blue & 1 & 10,000 & DN ($\times$10k) \\
NDVI & $-$10,000 & 10,000 & Index ($\times$10k) \\
NDWI & $-$10,000 & 10,000 & Index ($\times$10k) \\
NDBI & $-$10,000 & 10,000 & Index ($\times$10k) \\
Albedo & 1 & 9,980 & Ratio ($\times$10k) \\
\bottomrule
\end{tabular}
\end{table}

\begin{equation}
\label{ndvieq}
\text{NDVI} = \frac{\rho_{\text{NIR}} - \rho_{\text{Red}}}{\rho_{\text{NIR}} + \rho_{\text{Red}}}
\end{equation}
\begin{equation}
\label{ndwieq}
\text{NDWI} = \frac{\rho_{\text{Green}} - \rho_{\text{NIR}}}{\rho_{\text{Green}} + \rho_{\text{NIR}}}
\end{equation}
\begin{equation}
\label{NDBIeq}
\text{NDBI} = \frac{\rho_{\text{SWIR1}} - \rho_{\text{NIR}}}{\rho_{\text{SWIR1}} + \rho_{\text{NIR}}}
\end{equation}

\subsection{Local Climate Zone Labels}
For stratified evaluation across land-surface types, each pixel receives one of the 17 Local Climate Zone (LCZ) classes~\cite{stewart2012local} from the 2020 CONUS LCZ product~\cite{demuzere2020combining}. An 18th code represents custom or unclassified pixels. LCZ codes 1--10 cover urban classes, from compact high-rise to heavy industry. Codes 11--17 cover natural land-cover types, from dense trees to water. The HeatCast distribution is imbalanced (Table~\ref{tab:lcz_fractions}). Open urban classes (LCZ~4--6) cover 58.1\% of valid pixels, and natural classes (LCZ~11--17) cover 36.2\%. We therefore report LCZ-stratified errors in Section~\ref{sec:protocol}.

\begin{figure}[t]
\centering
\includegraphics[width=\linewidth]{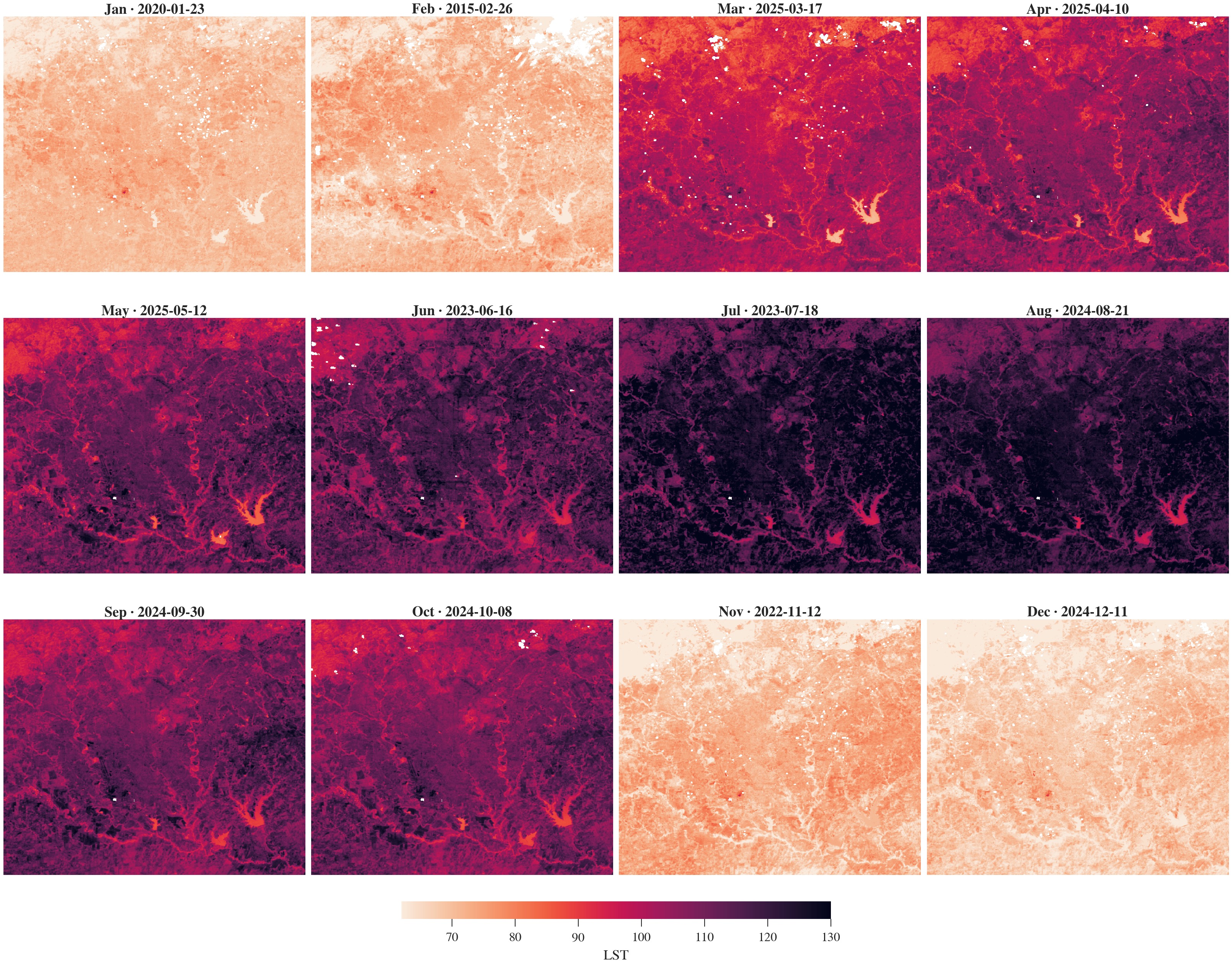}
\caption{\textbf{San Antonio shows a strong annual LST cycle.} Each panel uses the highest-coverage scene for that calendar month, preferring 2024 and using another year when needed. All panels share one color scale.}
\Description{A 3 by 4 grid of twelve San Antonio land-surface-temperature heatmaps, one for each calendar month, sharing a single colorbar, with cooler colors in winter and warmer colors in summer.}
\label{fig:san_antonio_cycle}
\end{figure}

\begin{table}[t]
\centering
\small
\caption{LCZ super-cluster coverage in HeatCast, computed across the CONUS LCZ product over the 124 city footprints. The remaining $\sim$1.4\% is custom/unclassified (LCZ~18).}
\label{tab:lcz_fractions}
\begin{tabular}{@{}llr@{}}
\toprule
\textbf{Super-cluster} & \textbf{LCZ codes} & \textbf{Pixels (\%)} \\
\midrule
Compact urban & 1--3 & 0.76 \\
Open urban & 4--6 & 58.13 \\
Other urban & 7--10 & 3.55 \\
Natural & 11--17 & 36.18 \\
\bottomrule
\end{tabular}
\end{table}

\subsection{Tiling, Packaging, and Statistics}
\begin{figure*}[t]
\centering
\includegraphics[width=0.95\textwidth]{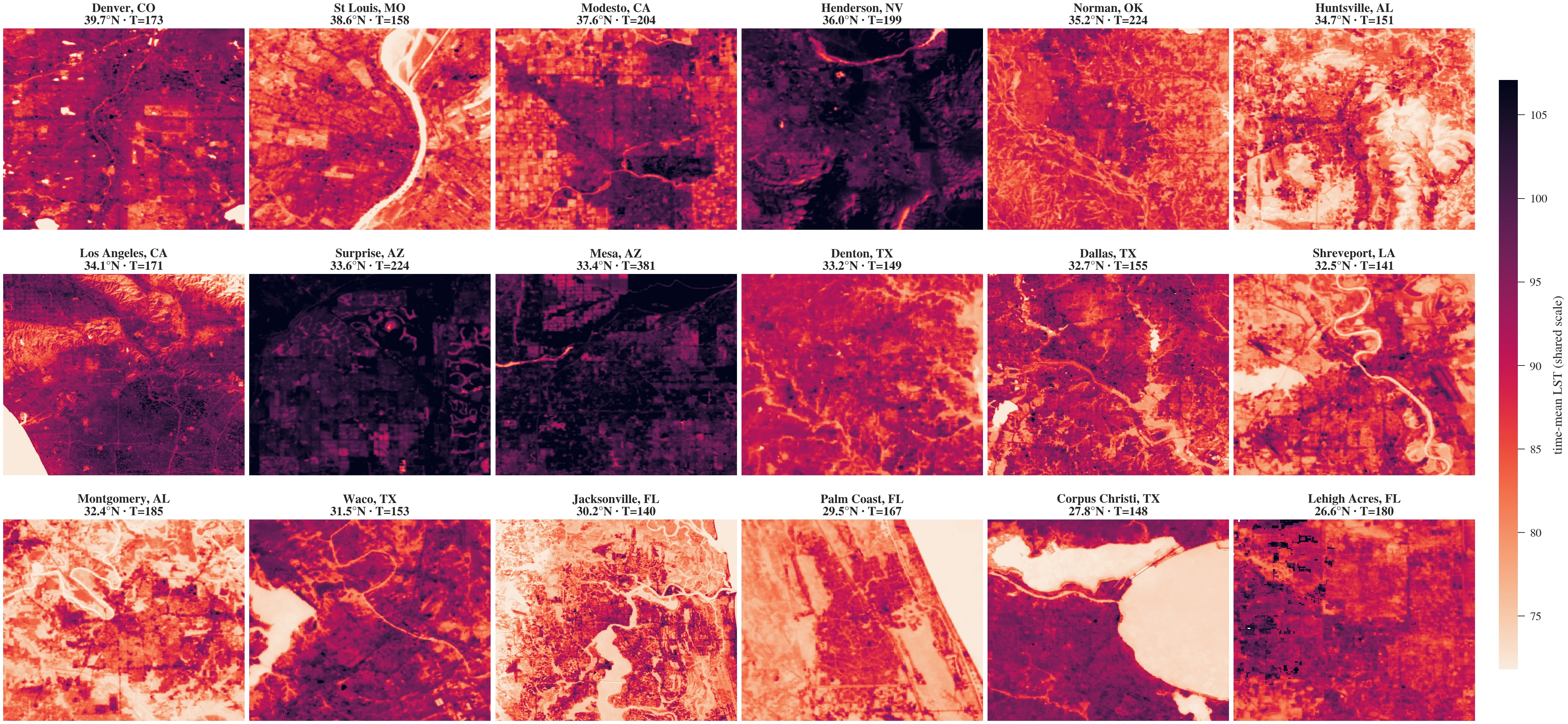}
\caption{\textbf{Intra-city LST patterns appear across the benchmark.} The panels show time-mean LST for eighteen cities sampled evenly across latitude and rendered with one color scale. Built corridors, parks, water bodies, and airports remain visible on the 30~m grid.}
\Description{A three-row by six-column grid of eighteen city heat-map panels sorted north-to-south, sharing a single LST colorbar on the right, with darker reds indicating warmer pixels.}
\label{fig:uhi_gallery}
\end{figure*}

We partition each city into individual Zarr~v3 stores containing the nine-channel monthly stack with non-overlapping $128 \times 128$~pixel patches. Each patch is $3.84$~km on a side and covers approximately 14.75~km$^2$. Each store has one array per variable, sharded chunks along $(T, Y, X)$, and Blosc compression. This layout supports streaming with HTTP range reads and variable stacking without rearranging data on disk. The DEM is static per pixel. Cloud-affected and invalid pixels remain explicit NoData and are propagated through the derived indices. Table~\ref{tab:dataset_stats} summarizes the dataset. Figures~\ref{fig:san_antonio_cycle} and~\ref{fig:uhi_gallery} show the monthly cadence and time-mean LST across cities.

\subsection{Quality Control}
We mark pixels as NoData before computing spectral indices when the Landsat QA\_PIXEL band flags cloud, cloud shadow, dilated cloud, or invalid data. Within an input sequence, we permit per-pixel linear interpolation across at most two consecutive missing months. We exclude tiles with longer gaps to keep cloud-gap reconstruction separate from the forecasting task. All evaluation code reports error in Kelvin and masks invalid pixels before aggregation. The release includes the masks used for the reported results so that future methods are evaluated on the same pixels.

\begin{table}[t]
\centering
\caption{HeatCast summary statistics. LST errors are evaluated and reported in Kelvin.}
\label{tab:dataset_stats}
\begin{tabular}{@{}ll@{}}
\toprule
\textbf{Attribute} & \textbf{Value} \\
\midrule
Cities & 124 (CONUS) \\
Temporal coverage & 2013 -- Jun~2025 \\
Temporal cadence & Monthly \\
Spatial resolution & 30~m \\
Tile size & $128 \times 128$ ($\sim$14.75 km$^2$) \\
Input channels & 9 \\
Output channel & 1 (LST) \\
Total tiles & $\sim$1.4~M \\
Storage & $\sim$150~GB \\
\bottomrule
\end{tabular}
\end{table}

\section{Benchmark Task and Evaluation Protocol}
\label{sec:protocol}

\subsection{Forecasting Task}
We define a single primary task: given $T$ consecutive monthly observations of a tile, predict the LST channel at the next month. Formally,
\begin{equation}
f_\theta: \mathbf{X} \in \mathbb{R}^{T \times C \times H \times W} \;\rightarrow\; \widehat{\mathbf{Y}} \in \mathbb{R}^{H \times W},
\end{equation}
with $T=12$, $C=9$, $H=W=128$. The 12-month context exposes models to a full seasonal pattern. Missing months within an input window are handled by the interpolation-and-exclusion rule of Section~\ref{sec:dataset}.

\subsection{Split}
\begin{figure*}[t]
\centering
\includegraphics[width=0.90\textwidth]{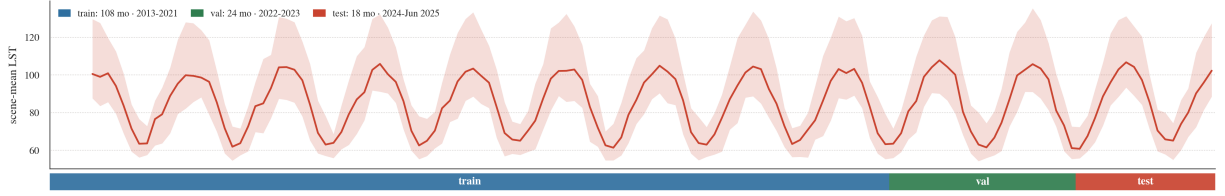}
\caption{\textbf{Temporal split overlaid on the benchmark-wide LST trace.} The line is the median per-city scene-mean LST across the 124 cities at each monthly step; the shaded band is the 10th--90th percentile envelope. The colored strip under the x-axis marks the partition: 108 nominal training months (Jan~2013--Dec~2021; the earliest Landsat~8 scenes begin mid-2013), 24 validation months (Jan~2022--Dec~2023), and 18 test months (Jan~2024--Jun~2025).}
\Description{A thin line plot of median per-city scene-mean LST from 2013 to mid-2025 with a 10-90 percentile envelope, plus a colored band along the bottom of the x-axis dividing the timeline into train, validation, and test partitions.}
\label{fig:splits_timeline}
\end{figure*}

We use a temporal split with training data from 2013--2021, validation data from 2022--2023, and test data from January~2024--June~2025. All 124 cities appear in each split, providing a chronological holdout across the benchmark's full geographic scope. The release includes fixed manifests for each split. Figure~\ref{fig:splits_timeline} overlays the split on the benchmark-wide LST trace.

\subsection{Metrics}
We compute RMSE (in Kelvin) over valid pixels within each tile and report the mean tile-level RMSE, giving each tile equal weight. We also report RMSE for four LCZ groups: compact urban (LCZ~1--3), open urban (LCZ~4--6), other urban (LCZ~7--10), and natural (LCZ~11--17). The reference evaluation script applies these definitions to all submissions.

\subsection{Design Rationale}
\begin{figure*}[ht!]
\centering
\includegraphics[width=0.75\textwidth]{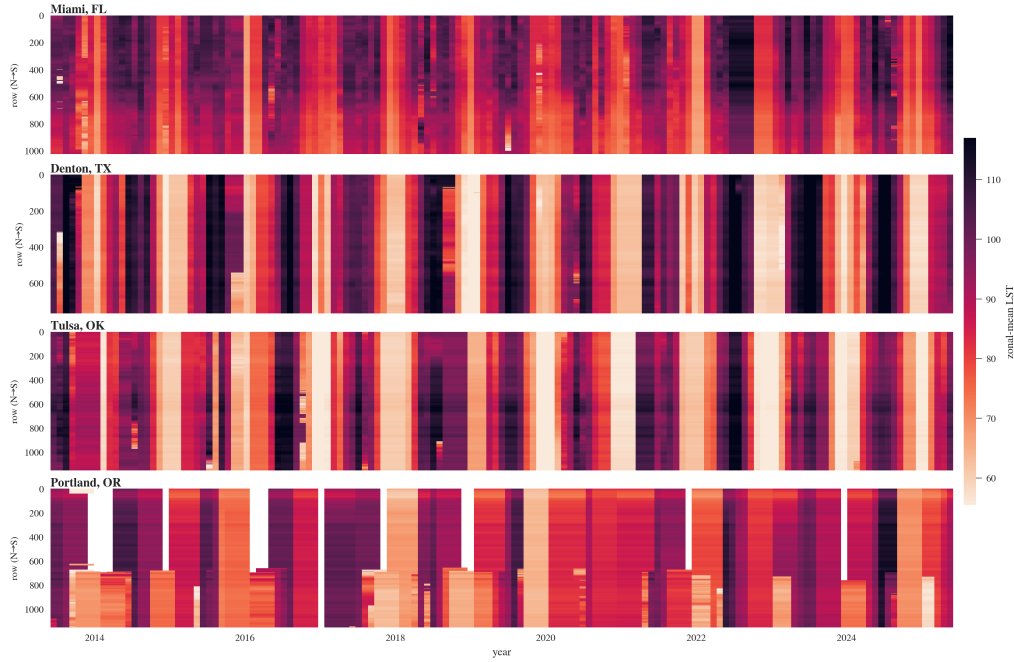}
\caption{\textbf{Seasonal and spatial LST patterns differ across cities.} Each Hovmöller diagram shows time horizontally, north-to-south position vertically, and zonal-mean LST by color.}
\Description{Four stacked panels, each showing a Hovmöller diagram for one city.}
\label{fig:hovmoller_regimes}
\end{figure*}

We choose a one-month forecast horizon for seasonal heat planning. A 12-month input window ensures the model captures a complete annual cycle, accounting for the varying intensity of seasonal peaks across different climates (Figure~\ref{fig:hovmoller_regimes}). We report LCZ-specific RMSE because aggregate RMSE is dominated by open urban and natural pixels.

\subsection{Quickstart}
\label{sec:quickstart}

The full $\sim$150\,GB release can be downloaded via \texttt{hf download -{}-repo-type dataset JesseGuerreroML/HeatCast}. A PyTorch wrapper then handles year splits, channel selection, NoData filtering, and sequence caching (Listing~\ref{lst:torch}).

\noindent\begin{minipage}{\linewidth}
\begin{lstlisting}[style=pythonstyle,caption={Loading HeatCast with the released \\ \texttt{LandsatSequenceDataset}: temporal split, 12-month input window, 1-month forecast horizon, and a \texttt{DataLoader} ready for training.},label={lst:torch},captionpos=b]
from torch.utils.data import DataLoader
from dataset import LandsatSequenceDataset

train_ds = LandsatSequenceDataset(
    dataset_root="path/to/HeatCast/",
    cluster="all",
    input_sequence_length=12,
    output_sequence_length=1,
    split="train",
    train_years=list(range(2013, 2022)),
    val_years=[2022, 2023],
    test_years=[2024, 2025],
    max_input_nodata_pct=0.60,
)

loader = DataLoader(train_ds, batch_size=32,
                    shuffle=True, num_workers=8,
                    pin_memory=True)

x, y = next(iter(loader))
# x: (B, 12, 9, 128, 128) -- 9 channels
# y: (B, 1,  1, 128, 128) -- next-month LST
\end{lstlisting}
\end{minipage}

Passing \texttt{remove\_channels=["LST"]} to the dataset reproduces the \emph{Auxiliary (no LST)} row of Table~\ref{tab:ablation_results}. The sequence cache rebuilds automatically.

\subsection{Release Contents}
The release contains city metadata, per-city Zarr~v3 stores, split manifests, baseline configurations and training scripts, pretrained checkpoints, and the reference evaluator. The CNN+LSTM and Earthformer checkpoints match Tables~\ref{tab:lstm} and~\ref{tab:ablation_results}. The evaluator computes aggregate, per-city, and LCZ-stratified RMSE. Patch identity is deterministic in $(\text{city}, y, m, row, col)$, so the reported numbers can be regenerated from the downloaded artifacts.

The dataset is versioned independently of the baseline code so that the benchmark remains usable as stronger baselines are added. Submissions should report the dataset version, code commit, random seeds, and any pretrained weights or external data used.

\begin{figure}[t]
\centering
\begin{lstlisting}[style=bashstyle,caption={},label={lst:zarr_tree},aboveskip=0pt,belowskip=0pt]
JesseGuerreroML/HeatCast/               # one Zarr v3 store per city
|-- <City_ST>.zarr/
|   |-- zarr.json                    # group metadata (attrs)
|   |-- LST/    (T, Y, X)  int16     # forecasting target
|   |-- albedo/ (T, Y, X)  int16
|   |-- blue/   green/   red/        # raw Landsat reflectance
|   |-- ndvi/   ndwi/    ndbi/       # spectral indices
|   `-- ...                          # all arrays share the 30 m grid
`-- analysis/heatcast_analysis.zip   # pre-computed per-city aggregates
\end{lstlisting}
\begin{lstlisting}[style=pythonstyle,aboveskip=0pt,belowskip=0pt]
import xarray as xr
ds = xr.open_zarr(
    "https://hf.co/datasets/JesseGuerreroML/HeatCast"
    "/resolve/main/San_Antonio_TX.zarr",
    consolidated=False, chunks={},
)
ds["LST"].isel(time=slice(0, 12)).load()
\end{lstlisting}
\caption{\textbf{Each city is distributed as a Zarr~v3 store.} Channels use sharded \texttt{(T, Y, X)} arrays that xarray can read from the Hub over HTTP without cloning the full dataset.}
\Description{A code listing showing the per-city Zarr v3 layout and a short xarray snippet that opens one city's store directly from the Hugging Face Hub.}
\label{fig:release_artifacts}
\end{figure}

\section{Baselines}
\label{sec:baselines}

We evaluate a CNN+LSTM~\cite{yao2018deep, hochreiter1997long} and Earthformer~\cite{gao2022earthformer} on the temporal split. We release both implementations, configurations, and checkpoints. The CNN+LSTM uses a two-layer convolutional encoder with 64 hidden channels and $3\times3$ kernels. A 128-unit LSTM models time, and a linear decoder produces a $128\times128$ prediction. Prior LST and urban-flow studies use this architecture~\cite{kartal2022prediction, maddu2021prediction}. Earthformer uses cuboid attention for spatio-temporal Earth-system forecasting. We use the \emph{earthnet} configuration with 256 base units, four heads, encoder and decoder depths $[1,1]$, cuboid size $(2,4,4)$, eight global vectors, and a stack-conv downsampler.

Both models take $12 \times 9 \times 128 \times 128$ inputs and use AdamW~\cite{loshchilov2019adamw} with a learning rate of $5\times10^{-4}$. We train in fp32 with batch size 32 for at most 200 epochs and use early stopping~\cite{10.5555/645754.668392}. Both models use random rotations and flips. The implementations use PyTorch~\cite{paszke2019pytorch}. Table~\ref{tab:baseline_config} summarizes the configurations, and the release contains the full files, seeds, and checkpoint paths.

\section{Experiments}
\label{sec:experiments}

\subsection{Baseline Results}
Table~\ref{tab:lstm} reports test-set RMSE in Kelvin on the temporal split. Earthformer obtains 7.74~K aggregate RMSE, compared with 10.42~K for the CNN+LSTM. Earthformer also has lower RMSE on open urban, other urban, and natural LCZ categories. The CNN+LSTM has lower RMSE on compact urban pixels (8.62~K versus 12.68~K), which account for 0.76\% of valid pixels.

\subsection{Feature-Set Ablation}
We evaluate the input channels by retraining each model on three configurations. \emph{LST only} uses 12 months of historical LST. \emph{Auxiliary (no LST)} uses the eight non-LST channels for 12 months. \emph{RGB only} uses the three surface-reflectance RGB channels for 12 months.

Earthformer obtains its lowest RMSE with the auxiliary channels (7.72~K), followed by all channels (7.74~K) and historical LST alone (8.15~K). Using only auxiliary channels improves RMSE by 0.43 K compared to using LST history alone. For the CNN+LSTM, the full configuration is best, and removing historical LST increases RMSE by 2.05~K.

\begin{table}[t]
\centering
\caption{\textbf{Baseline results.} Test-set RMSE (K, $\downarrow$) on the temporal split, stratified by LCZ super-cluster. Lower is better; best value per row in \textbf{bold}.}
\label{tab:lstm}
\begin{tabular}{@{}lcc@{}}
\toprule
\textbf{Subset} & \textbf{CNN+LSTM} & \textbf{Earthformer} \\
\midrule
All LCZs & 10.42 & \best{7.74} \\
Compact urban & \best{8.62} & 12.68 \\
Open urban & 10.82 & \best{8.41} \\
Other urban & 8.01 & \best{6.99} \\
Natural & 9.78 & \best{6.58} \\
\bottomrule
\end{tabular}
\end{table}

\begin{table}[h]
\centering
\small
\caption{Baseline configurations. Both models use AdamW, batch size 32, fp32, at most 200 epochs with early stopping, and $12\times9\times128\times128$ inputs.}
\label{tab:baseline_config}
\begin{tabular}{@{}lll@{}}
\toprule
 & \textbf{CNN+LSTM} & \textbf{Earthformer} \\
\midrule
Hidden / base & 64 conv, 128 LSTM & 256 base units \\
Depth & 2 conv layers & enc/dec $[1,1]$ \\
Attention & --- & 4 heads, cuboid $(2,4,4)$ \\
Global vectors & --- & 8 \\
Downsampler & --- & stack-conv $[64,256]$ \\
Learning rate & $5\times10^{-4}$ & $5\times10^{-4}$ \\
\bottomrule
\end{tabular}
\end{table}

\begin{table}[t]
\centering
\caption{\textbf{Earthformer achieves its lowest RMSE with auxiliary inputs.} Test-set RMSE (K, $\downarrow$) for the feature-set ablation. The best value for each model is in \textbf{bold}.}
\label{tab:ablation_results}
\begin{tabular}{lcc}
\toprule
\textbf{Input configuration} & \textbf{CNN+LSTM} & \textbf{Earthformer} \\
\midrule
All channels (default)  & \best{10.42} & 7.74 \\
LST only                & 11.09 & 8.15 \\
Auxiliary (no LST)      & 12.47 & \best{7.72} \\
RGB only                & 14.82 & 8.68 \\
\bottomrule
\end{tabular}
\end{table}

\subsection{Reproducibility}
The reported numbers can be regenerated from the released manifests, configurations, and checkpoints. The reference evaluator applies the fixed NoData masks and converts errors to Kelvin. The release also includes a small smoke-test subset of selected cities and months so that users can verify their environment without downloading the 150~GB archive.

\begin{figure}[t]
\centering
\includegraphics[width=\linewidth]{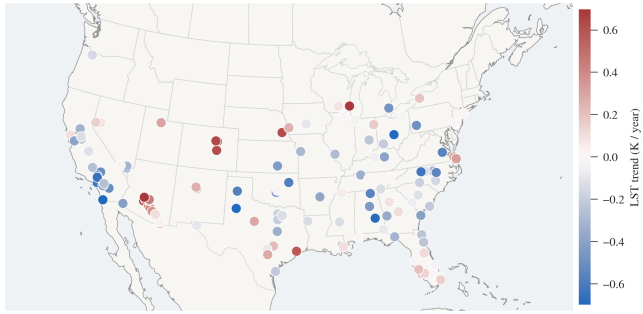}
\caption{\textbf{Annual-mean LST trends vary across HeatCast cities.} The map shows the OLS slope in K per year for cities with at least six years of QC-passed observations.}
\Description{An Albers map of the United States with one colored dot per HeatCast city, encoding the per-city LST trend in Kelvin per year.}
\label{fig:warming_map}
\end{figure}

\begin{figure}[t]
\centering
\includegraphics[width=0.55\linewidth]{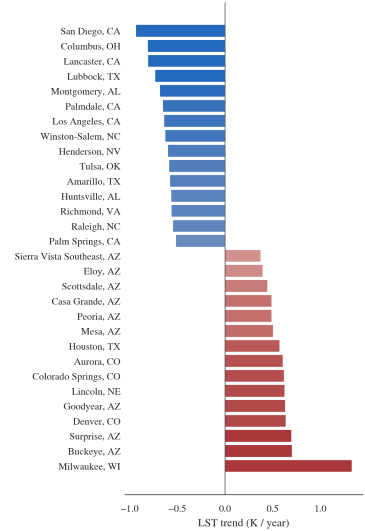}
\caption{\textbf{Ranked LST trend extremes.} The fifteen most strongly cooling and fifteen most strongly warming cities in HeatCast, ordered by OLS slope of annual-mean LST.}
\Description{A horizontal bar chart of the top fifteen cooling and top fifteen warming cities, sorted by LST trend.}
\label{fig:warming_bar}
\end{figure}

Figures~\ref{fig:warming_map} and~\ref{fig:warming_bar} provide an additional descriptive analysis of per-city LST trends using the released data.

\section{Research Directions}
\label{sec:directions}
HeatCast supports extensions in model comparison, feature fusion, and class-balanced forecasting.

\noindent\textbf{Seasonal reference models.} The fixed manifests and evaluator support direct comparisons with persistence and monthly climatology. These reference models can quantify the predictive gain beyond the annual LST cycle.

\noindent\textbf{Feature fusion.} Auxiliary inputs produce the lowest Earthformer RMSE in Table~\ref{tab:ablation_results}. Multi-stream encoders~\cite{simonyan2014two}, physically informed objectives~\cite{intracityPINNs}, and Landsat-pretrained models~\cite{stewart2023ssl4eo} provide direct ways to extend this result.

\noindent\textbf{Rare urban classes.} The CNN+LSTM outperforms Earthformer by 4.06~K on compact urban pixels (LCZ~1--3). Balanced sampling, class weighting, or conditioning on land-cover type may improve performance on these rare classes.

\section{Use Cases}
HeatCast can be joined with demographic boundaries and historical redlining maps for observational studies of urban heat. Prior work has linked Home Owners' Loan Corporation redlining grades to present-day LST differences in U.S. cities~\cite{hoffman2020effects, hashemi2024thermal}. Such studies require external demographic data and causal assumptions beyond this benchmark.

The aligned multispectral and thermal archive may also support pretraining and transfer to related Landsat tasks. Existing geospatial foundation models have not been evaluated widely on thermal forecasting~\cite{stewart2023ssl4eo, cong2022satmae, jakubik2023prithvi}.

\section{Limitations and Ethical Considerations}
HeatCast inherits uncertainty and coverage limits from its source products. Landsat Level-2 LST has 1--2~K uncertainty under cloud-free conditions and 100~m native support, although HeatCast aligns all channels to a 30~m grid~\cite{malakar2018landsat}. Each baseline uses one training run, so the 0.02~K difference between the two best Earthformer variants is not a stable ranking. The temporal split evaluates future dates in known cities rather than transfer to unseen cities. Coverage favors large CONUS cities with sufficient cloud-free imagery, and LCZ labels inherit source-product errors~\cite{demuzere2020combining}. LST measures surface rather than air temperature. The dataset contains no personal identifiers, but combining 30~m pixels with parcel or demographic data may raise privacy and equity concerns.

\paragraph{Data and code availability.}
{\sloppy
The HeatCast dataset (124 cities, 30~m, 2013--June~2025, $\sim$150~GB) is released under MIT at \url{https://doi.org/10.57967/hf/9889}. Code, configurations, the reference evaluator, and pretrained CNN+LSTM and Earthformer checkpoints matching Tables~\ref{tab:lstm} and~\ref{tab:ablation_results} are also released under MIT.
\par}

\section{Conclusion}
HeatCast defines monthly LST forecasting at 30~m across 124 U.S. cities. Earthformer improves aggregate RMSE from 10.42~K for the CNN+LSTM to 7.74~K. The auxiliary-only Earthformer obtains the best ablation RMSE at 7.72~K, showing that the non-LST channels provide strong predictive signal. The data, code, and weights are released under MIT at \url{https://doi.org/10.57967/hf/9889}.

\FloatBarrier
\section*{Acknowledgments}
The UTSA HPC platform provided compute. We used Claude Opus 4.7 (Anthropic) to review prose for clarity and to assist with figure-generation code. The authors verified the final text, figures, and results.

\bibliographystyle{ACM-Reference-Format}
\bibliography{refs}

\end{document}